\documentclass[runningheads]{llncs}
\usepackage{amsmath,amssymb,amsfonts}
\usepackage{booktabs} 
\usepackage{caption} 
\usepackage{subcaption} 
\usepackage{graphicx}
\usepackage{pgfplots}
\usepackage[utf8]{inputenc}
\usepackage{algpseudocode,algorithm,algorithmicx}
\usepackage{graphicx}
\newcommand{\R}{\mathbb{R}}

\newcommand{\cC}{\mathcal{C}}

\newcommand{\cO}{\mathcal{O}}

\newcommand{\cZ}{\mathcal{Z}}
\newcommand{\cT}{\mathcal{T}}
\newcommand{\cX}{\mathcal{X}}

\newcommand{\la}{\lambda}
\newcommand{\ga}{\gamma}

\newcommand{\bv}{\mathbf{v}}

\newcommand{\bu}{\mathbf{u}}
\newcommand{\bw}{\mathbf{w}}
\newcommand{\hbw}{\mathbf{\hat{w}}}

\newcommand{\bx}{\mathbf{x}}

\newcommand{ \simi }{{\rm sim}}

\definecolor{blue}{HTML}{1F77B4}
\definecolor{orange}{HTML}{FF7F0E}
\definecolor{green}{HTML}{2CA02C}

\pgfplotsset{compat=1.14}

\begin{document}
\title{Multiway clustering of 3-order tensor via affinity matrix}
%
%
\author{Dina Faneva Andriantsiory\inst{1} \and
Joseph Ben Geloun\inst{1} \and
Mustapha Lebbah\inst{2}}
%
%
\institute{   LIPN, UMR CNRS 7030  , Sorbonne Paris Nord University, Villetaneuse, France \\ 
\and
DAVID Lab, University of Versailles, Universit\'e Paris-Saclay, Versailles, France 
}
\maketitle              

\begin{abstract}
 We propose a new method of multiway clustering for 3-order tensors via affinity matrix (MCAM). Based on a notion of similarity between the tensor slices and the spread of information of each slice, our model builds an affinity/similarity matrix on which we apply advanced clustering methods. The combination of all clusters of the three modes delivers the desired multiway clustering. Finally, MCAM achieves competitive results compared with other known algorithms on synthetics and real datasets.

 \end{abstract}

\section{Introduction}
\label{sec:introduction}
Tensor data is seen as multidimensional arrays that
structure much complex information. 
This occurs in a variety of domains such as time-evolving data \cite{araujo2018tensorcast}, behavioral patterns \cite{he2018inferring}, heterogeneous information networks \cite{ermics2015link}, and social networks \cite{nickel2011three}. 
Developing algorithms to understand the characteristics of different patterns in this data type remains a considerable challenge for data scientists.  Machine learning methods define powerful statistical tools to undertake the mining of multidimensional data \cite{cichocki2009nonnegative}.
Sundry algebraic tools were developed to gather the information 
that lies in a subspace of the tensor dataset 
such as the representation of a tensor with a finite sum of rank-one decomposition \cite{Harshman1996CPTuckerDecomposition,kiersCP2000towards,kolda2009tensor}. 
Alternatively, one can reduce the initial tensor to a compressed tensor, i.e. to a tensor with a smaller size called \textbf{core tensor}, 
which leads to a different approach to understanding the dataset \cite{Harshman1996CPTuckerDecomposition,tucker1966}.

One of the most prominent approaches to tensor pattern recognition  declines in clustering algorithms.  In particular, clustering which is unsupervised learning 
 attracts substantial attention because of its considerable range of applications. Clustering refers to the partition of data into clusters (groups) of similar objects. Each cluster consists of objects that are similar to each other and dissimilar to objects in other groups \cite{berkhin2006survey}.

Clustering algorithms for three-dimensional datasets are well-developed nowadays. In the following, we give a lightening review of two different types of clustering methods for tensor data: some methods require the number of clusters and  some others do not. The Tucker+k-means method \cite{HuangHOOISVD_kmeans}  starts with the Tucker decomposition \cite{tucker1966,kolda2009tensor} and leads to the core tensor with the three membership matrices. Independent of each other, the membership matrices from the tensor decomposition keep the variation of the data in each mode. Then the k-means algorithm \cite{hartigan1979algorithm_kmeans} 
applies to each membership matrix to determine the elements of the clusters in each mode.

The CANDECOMP/PARAFAC decomposition of a tensor \cite{kiersCP2000towards} with k-means
(CP+k-means) 
is regarded as a particular case of the Tucker+k-means method. Indeed, in the CP decomposition, the core tensor is structured into a superdiagonal 
three-way tensor (for a cubical 3-order tensor, this is the body diagonal of the
tensor) and the data is represented as a sum of component rank-one tensors which are the columns of the membership matrices \cite{kiersCP2000towards,kolda2009tensor}. Once again, we run a k-means to each membership matrix to determine the different clusters in a different  mode.

Another key clustering method is 
the so-called multiway clustering via tensor block models (TBM) \cite{WangMultiwayClustering2019}. 
This approach uses a particular version of the Tucker decomposition: in each mode, the corresponding membership matrix that contains only 0 and 1, becomes the clustering partition. 
The heterogeneous tensor decomposition for clustering via manifold optimization \cite{YSun2016Manifold} is also based on a specific Tucker decomposition called the heterogeneous model. The clustering membership information is recorded in the last membership matrix. This matrix gets updated over the multinomial manifold principle. 

All the previous clustering methods take as a hyper-parameter the number of clusters in each mode. Other algorithms do not require such a hyper-parameter. 
Among those, we cite the Parameter-Less Tensor Co-clustering \cite{BattagliaParameterLess2019}. This algorithm is adapted to the non-negative tensors and is based on Goodman-Kruskal's $\tau$ association measure \cite{goodman1979measures}. It maximizes the scalarization function with a stochastical local search to find the different clusters in all modes.
The new input parameter needed for this algorithm is the
number of iterations. 
We could also mention some algorithms that extract a unique cluster that contains the most relevant information satisfying some criterion of similarity. Among those, the Tensor Biclustering algorithm \cite{FeiziNIPS2017} determines the highly correlated trajectories over the third dimension that lies in a subspace (rank-one tensor). To apply this scheme, 
one needs the size of the cluster in each mode
as an input. 
More recently, using neither the number of clusters
nor the cluster size, the multi-slice clustering algorithm \cite{andriantsiory2021multislice} is based on the comparison of the spread of information for each tensor slice. This algorithm builds an affinity or similarity matrix between slices for each tensor mode and uses a threshold error to guarantee the similarity quality of the output cluster. 

\begin{figure}[ht]
\includegraphics[width=5cm, height=4.5cm]{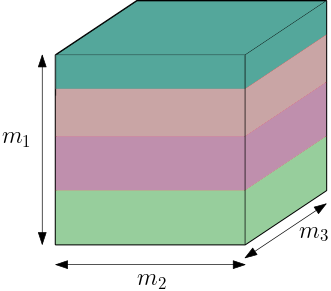}
\hfill
\includegraphics[width=5cm, height=4.5cm]{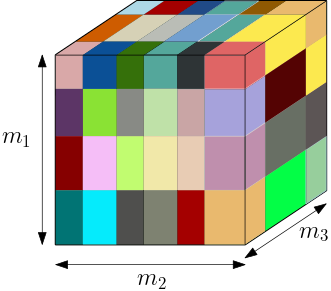}	
\caption{Mode-1 clustering (left) and multiway clustering (right) of a 3-order tensor.} 	 	
\label{fig:multiway}
\end{figure}

Our contribution in this paper is to present new approaches for multiway clustering for $3$-order tensors based on the affinity (similarity) matrix inspired by
\cite{andriantsiory2021multislice}: 
in a given mode,  each entry of the matrix represents the similarity of two slices 
in this mode. However, departing from the method proposed in the above reference that singles out a unique cluster, we undertake the analysis of this affinity matrix using advanced multiway-clustering algorithms based on spectral analysis \cite{BachLearningSpectralClustering} and affinity propagation \cite{frey2007clustering}. These methods are particularly appropriate in our situation
as they are designed for affinity matrices. 
We titled our method Multiway Clustering via Affinity Matrix (MCAM) where the algorithms determine the clusters in each mode independently. 
The combination of all clusters of the three modes provides the multiway clustering as illustrated in figure \ref{fig:multiway}.
Another benefit of the present method is its adaptability in two situations depending on whether the number of clusters is given as input or not.
 Hence, the users may choose
among these two options. 
In addition, the MCAM is portable in the sense that one may choose any other matrix clustering algorithm to perform the clustering on its affinity matrix. In this work, 
as mentioned above, we focus on two well-established algorithms, spectral clustering, and affinity propagation. 
The MCAM performs well on both synthetic and real datasets. 
We compare its results with three known
and well-performing clustering algorithms for tensorial data, namely 
Tucker+k-means, CP+k-means, and TBM. 
We find that the MCAM is efficient with a few numbers of the largest eigenvalues 
with their corresponding eigenvectors of the slice covariance matrices,
whereas the other methods require a much higher number of rank-one tensors in the decomposition of the tensor dataset
before becoming truly efficient. This, therefore, privileges
the MCAM for generic tensor data that do not assume any particular form of the tensor data. 

 We present two variant algorithms of the MCAM. Both algorithms are competitive and efficient compared to other methods. The first algorithm called MCAM-I has a domain of validity for any generic tensor dataset. The second algorithm, which we call MCAM-II, works and performs  better than MCAM-I in some particular situations where no correlation occurs
between leading and subleading eigenvalue-eigenvector pairs of different slice covariance matrices. We find it useful to report both results for both algorithms as this already allows us to gauge the clustering efficiency of MCAM.

The structure of this paper is the following. Section \ref{sec:methodology} and section \ref{sec:algorithms} present respectively the methodology of MCAM and derive two algorithms. 
For the reproducibility purpose, we open-sourced our experimental setup and model implementations at
 the link \url{https://github.com/ANDRIANTSIORY/MCAM}.
Section \ref{sec:experimentation} discusses some experiments
on synthetic and real datasets. With the results, we conclude  that the MCAM is a valid
multiway cluster detector. We also compare the performance of MCAM  with three known  algorithms (CP+k-means,  Tucker+k-means and TBM) before summarizing our results in a conclusion in section  \ref{sec:conclusion}. 

\section{Methodology}

\label{sec:methodology}

\subsection{Notation }
We denote by $\cT, \cX, \cZ$ the tensor dataset, the signal tensor, and the noise tensor, respectively.  Matrices are represented by capital letters ($T, X, Z,\cdots$). For a matrix $V$, $|V|$ denotes a matrix with the same size as $V$ and its entries are the absolute values of the corresponding entries in $V$. $V^t$ represents the transpose of the matrix $V$. Vectors and real numbers are denoted by  boldface lowercase letters and lowercase letters,  respectively. For a nonzero integer $n$, $[n]= \{1,2,\cdots,n\}$. For a matrix, the largest eigenvalue is called the \textbf{top eigenvalue}, 
and its corresponding eigenvector 
\textbf{top eigenvector}. 


We consider a tensor $\cT = \cX + \cZ$ with $\cT\in\R^{m_1\times m_2\times m_3}$, where $\cX$ is the signal tensor, and $\cZ$ is the noise tensor.
The CP decomposition \cite{kolda2009tensor} consists in writing the tensor in the form of a finite sum of the rank-one tensors. It can be viewed as a Tucker decomposition where the core tensor is super-diagonal. Hence, for a signal tensor $\cX$, its CP decomposition with $d$ rank-one tensors
($d$ a positive integer) finds the form 
\begin{equation}
    \cX \approx \sum_{j=1}^{d} \ga_j \bu_j\otimes \bv_j\otimes\bw_j,
\end{equation}
then $\cT$ can be written as
\begin{equation}\label{eq:CPdecomposition}
    \cT \approx \sum_{j=1}^{d} \ga_j \bu_j\otimes \bv_j\otimes\bw_j + \cZ ,
\end{equation}
where $\bu_j\in\R^{m_1}, \bv_j\in \R^{m_2}, \bw_j\in \R^{m_3}, \|\bu_j\|=\|\bv_j\|=\|\bw_j\|=1,$ 
for $\ga_j\in\R_+^*$ for all $j\in \{1,\cdots, d\}$ and $d \ll \min(m_1, m_2, m_3)$,  $\ga_j$ is the weight of each  rank-one component of the tensor.

We assume that the entries of the noise tensor $\cZ$ are independent identically distributed (i.i.d) standard normal random variables. 

Given a mode, the matrix that records the slice similarity is called the affinity or similarity matrix. The idea of constructing  the matrix representing the slice similarity in each mode of the tensor has been introduced in  \cite{andriantsiory2021multislice}
and has led to the so-called 
the multi-slice clustering for a rank-one 3-order tensor. This algorithm delivers a cluster of slices based on a threshold parameter
that gauges the similarity between slices. We bring an improvement to this method to determine the multiway clustering of the 3-order tensor.
 Note that, finally, there is no straightforward
comparison between the resulting clustering 
algorithms of the two methods: the multi-slice clustering
delivers a single cluster of slices in each mode of the tensor and no guarantee of the similarity of the remaining dataset, 
whereas MCAM delivers a partition of the slices in each mode, from which we identify several clusters with guaranteed similarity.

The strategy of the MCAM is declined in two phases. 
Firstly, we construct  the similarity matrix for a rank-one tensor dataset. We  generalize this construction of the similarity matrix for any tensor dataset.
 Secondly, once the similarity matrix is obtained, the next task  is to apply an advanced matrix clustering to deliver the multiway clustering. 
We will focus on the so-called spectral clustering method \cite{BachLearningSpectralClustering} and the affinity propagation  \cite{frey2007clustering} 
to extract the set of clusters in each mode. 
Note that the following explanations concentrate on the mode-$1$ of the tensor and there is no issue with extending the reasoning for the remaining modes.

\subsection{Affinity matrix for a rank-one tensor (equation \eqref{eq:CPdecomposition} with $d=1$)} We use the Matlab notation for the manipulation of the tensor. In mode-1 of the tensor $\cT$, we have $m_1$ slices, and the $i$-th slice is defined by:
\begin{equation}  \label{eq:slice}
T_i = \cT(i,:,:)=\cX(i,:,:) + \cZ(i,:,:),\qquad \forall i\in[m_1]
\end{equation}
where $\cX(i,:,:)$ and $\cZ(i,:,:)$  are respectively the $i$-th slice of the signal tensor and the $i$-th slice of the noise tensor, and $T_i$ is a matrix in $\R^{m_2\times m_3}$. The principal component of the columns of $T_i$ is the largest eigenvector of $T_i^tT_i$ \cite{shlens2014tutorialPCA}.
The covariance matrix of $T_i$ is expressed as the following:
\begin{equation}
    T_i^t T_i =  \hat\ga_1^{2} \bw_1\bw_1^{t} + W_i,
    \label{eq:covarianceTi}
\end{equation}
where $\hat\ga_1 = \ga_1 \bu_1(i)$,\\
and
$W_i = X_i^tZ_i + Z_i^t X_i + Z_i^tZ_i $.

The first term of equation \eqref{eq:covarianceTi} represents the covariance matrix of the signal slice $X_i$ and $W_i$ refers to the noise. The mismatch between the subspace of the spectral decomposition of the slice $T_i^tT_i$ and the signal slice $X_i^tX_i$ is bounded (see Davis-Kahan $\sin\theta$ theorem \cite{yu2014useful}). 
For this noise model, the relationship 
between the top eigenvector of the covariance matrix~of the~$i$-th~slice and the corresponding signal is also controlled by the following proposition (see lemma 2,  supplementary material of \cite{andriantsiory2021multislice}).

\begin{proposition}
\label{lemCvv}
With the standard Gaussian noise model, let $\hbw_1$ be the top eigenvector of $T_i^tT_i\in \R^{m_3\times m_3}$, 
and 
   $\hat\ga_1 = \cO(m_3)$, and  $\alpha = \|\bw_1\|_{\infty}$. We have 
\begin{equation}
\label{eq:norm_infty}
	\|\hbw_1 - \bw_1\|_{\infty}\le \cO\left(\frac{1}{\hat\ga_1} \alpha\log(m_3)\right)
\end{equation}
with high probability as $m_3\rightarrow \infty$.
\end{proposition}

The equation \eqref{eq:norm_infty} indicates that, for a large $m_3$, if the signal weight becomes large then  the two vectors $\hat\bw_i$ and $ \bw_i$ become more similar. 

Each slice is represented by its top eigenvalue and top eigenvector. These provide the direction in which the data has maximum variance and in which the data is most spread out.
If one deals with a noise slice, we can determine the variance or top eigenvalue associated with its top eigenvector. 
An estimate of such a quantity is possible as the covariance matrix of a random standard Gaussian matrix, it has a white Wishart distribution \cite{johnstone2001}. 
We can therefore approximate the top eigenvalue distribution of the covariance matrix by the Tracy-Widom distribution \cite{CHIANI201469}.  

In mode-1, let's
consider $V$ in $\R^{m_3\times m_1}$ a matrix with the column vector 
$\bx_i = (\la_i/\la) \times \hbw_i$, where  $\la = \max_{i\in[m_1]} \la_i$, $\la_i$ is the top eigenvalue and $\hbw_i$ is the top eigenvector of the  $i$-th slice covariance matrix $T_i^tT_i$ for $i\in[m_1]$. From now on, we drop the hat from the notation of  $\hbw_i$, as there will be no  possible confusion. 

The similarity matrix, called $C'$, of all slices associated with the mode-$1$ is defined as: 
\begin{equation}
    C' = |V^t V|
    \label{similarity1}
\end{equation}
where the similarity  of the slice $i$ and slice $j$  $$C'_{ij} =  \simi (\bx_i, \bx_j) = \sigma_i\sigma_j | \langle \bw_i, \bw_j \rangle |$$
and where $0\le \simi(\bx_i, \bx_j) \le 1$ and $\sigma_i = \la_i/\la$ for all $i$.

The slices $i$ and $j$ are 
similar if  $C'_{ij}$ is close to $1$, and they are dissimilar if  $C'_{ij}$ is close to zero.

\subsection{Affinity matrix for general 3-order tensor}
We assume that our tensor dataset is approximated by a signal of $d$ rank-one tensors plus a noise term. 
Building the similarity matrix by only 
considering  the top eigenvalue and top eigenvector of each slice may result in a loss 
of information for each slice because the exact rank-one decomposition of the tensor is unknown most of the time. To solve this issue, we propose a new construction method of the similarity matrix $C'$ that takes into account multiple eigenvalues and eigenvectors for each slice.

To illustrate the new method, namely the MCAM, we assume that each slice is represented by their $r$ largest eigenvalues and
their corresponding eigenvectors, $r\le d$. 
For simplicity, let's limit to the case
of two slices. The generic 
case can be inferred easily from this point. 
The ideal situation for two slices to be similar is that the $r$ eigenvalues and eigenvectors of the two matrices are pairwise identical. Without loss of generality, we take $r=2$ for the following explanation and
the computation of the similarity matrix. Once more, the case
$r\ge 2$ can be recovered without difficulties
as it involves all possible eigen-subspace pairs.

For fixed $i$-th and $j$-th slices, we select 
the eigenspace $(\bx_1^{(i)}, \bx_2^{(i)})$ and $(\bx_1^{(j)}, \bx_2^{(j)})$, respectively, such that 
\begin{equation}
\label{baxi}
\bx_a^{(i)} = (\la_a^{(i)}/\la)  \bw_a^{(i)},
\quad a \in\{1,2\}
\end{equation}
where  $\la$ keeps its meaning 
as $\max_{i\in [m_1]} \la_1^{(i)}$.\\

By definition of the eigenspace, we have 
 $\la_1^{(i)} \ge\la_2^{(i)}$
 and $\la_1^{(j)} \ge\la_2^{(j)}$. 
 The question here is: given a strong (resp. weak)
 correlation $|\langle \bx_1^{(i)},\bx_1^{(j)}\rangle|\sim 1$ (resp. $\sim 0$), what does this entail on the 
 cross similarities $|\langle \bx_1^{(i)},\bx_2^{(j)}\rangle|$
  and $|\langle \bx_2^{(i)},\bx_1^{(j)}\rangle|$?
  Note that there are two ways to obtain 
  $|\langle \bx,\bx'\rangle|\sim 0$: 
  either $\bx \perp \bx'$
  or the norm of one of these vectors gets close
  to 0. 
To answer the above question, we conduct the following case
study: 
\begin{itemize}
    \item Assume that $|\langle \bx_1^{(i)},\bx_1^{(j)}\rangle|$ is close to one, then the cross similarities $|\langle \bx_1^{(i)},\bx_2^{(j)}\rangle|$ and $|\langle \bx_2^{(i)},\bx_1^{(j)}\rangle|$ are close to zero due to the orthogonality of the vectors $\bx_1^{(i)}\perp \bx_2^{(i)}$ and $\bx_1^{(j)}\perp \bx_2^{(j)}$. 
    
    \item Assume that $|\langle \bx_1^{(i)},\bx_1^{(j)}\rangle|$ is close to zero. In this case, two possibilities may occur: 
    \begin{itemize}
        \item If $|\langle \bx_2^{(i)},\bx_2^{(j)}\rangle|$ close to one, then  the cross similarities $|\langle \bx_1^{(i)},\bx_2^{(j)}\rangle|$ and $|\langle \bx_2^{(i)},\bx_1^{(j)}\rangle|$ become close to zero because of the  orthogonality of the vectors in the slice eigen-subspace. 
        
        \item If $|\langle \bx_2^{(i)},\bx_2^{(j)}\rangle|$ close to zero, then we need to push further the analysis. 
        We have the following table, where the 
        choice of $i,j\in \{1,2\}$ holds without loss of generality:  
        \begin{equation}
        \begin{array}{|c|c|c|}
        \hline
         &  \bx_1^{(i)} \perp \bx_1^{(j)}  & \la_1^{(i)}/\la \sim 0\\
         \hline
         \bx_2^{(i)} \perp \bx_2^{(j)} & ? &\sqrt{} \\
        \hline
        \la_2^{(i)}/\la \sim 0 & ? & \sqrt{} \\ 
        \hline
        \end{array}
        \label{check}
        \end{equation}
       where the  symbol $\sqrt{}$ indicates the cases where the cross
       similarities $|\langle \bx_1^{(i)},\bx_2^{(j)}\rangle|$ and $|\langle \bx_2^{(i)},\bx_1^{(j)}\rangle|$ become close to 0. In the other cases, no conclusion can be reached. 

    \end{itemize}

\end{itemize}


The above case analysis reveals that, in some situations, the cross 
scalar products $|\langle \bx_k^{(i)},\bx_{k'}^{(j)}\rangle|$
may become close to 1 and can be non-negligible. 
In general, they need 
to be incorporated into the analysis
of the affinity matrix. 
However, in the following and some cases, the experiments show that their presence may result in less good clustering performance. 
This deserves a better understanding
and will be left for future research. 
Taking this into account, 
we implement two variant
algorithms.  Regarded valid to a full extent, one algorithm (MCAM-I) includes the cross scalar products
in the affinity matrix. 
The second algorithm (MCAM-II) 
neglects these cross terms. 
We will perform our analysis in parallel keeping track 
of the performance of both 
algorithms. 

For the algorithm MCAM-I, the matrix is a linear combination of matrices $C_{kk'}$ for $k, k'= 1,\dots, r$,  with entry $(C_{kk'})_{ij} =  |\langle \bx_k^{(i)},\bx_{k'}^{(j)}\rangle|$.
Then, we obtain $r^2$ matrices ($C_{11}, C_{12},\cdots,C_{rr}$) from  the eigen-subspaces made of the $r$ top eigenvectors of each slice. All matrices $(C_{kk'})_{k,k'\in[r]}$ are combined to build the similarity matrix $C'$ defined as:
 \begin{equation}\label{eq:Cg}
   C' = \frac{\la^2}{(\sum_{k=1}^r\la_{k})^2}\sum_{k=1}^{r}\sum_{k'=1}^{r} C_{kk'}
\end{equation}
On the other hand, MCAM-II proposes to take the linear combination of the matrices $C_{k}:= C_{kk}$ for $k=1,\cdots,r$. Then we obtain $r$ matrices $(C_{11},C_{22},\cdots,C_{rr})$ from the eigen-subspaces made of r top eigenvectors of each slice. The similarity matrix $C'$ defined as :
\begin{equation}
\label{eq:Cg2}
       C' = \frac{\la^2}{\sum_{k=1}^r\la_{k}^2}\sum_{k=1}^{r} C_{kk},
\end{equation}
In both methods, $\la_k = \max_{i} \la_k^{(i)}$ (in this sense $\la = \la_1$ 
of equation \eqref{baxi}).

 If the entry $C'_{ij}$ of the matrix $C'$ is close to one, this means that the $i$-th and $j$-th slices are similar.  Otherwise, $C'_{ij}$ close to zero indicates that the two slices are dissimilar. 

 We realize that the computational cost of the construction of similarity matrix of MCAM-I dominates that of the matrix of MCAM-II. However, our experiments show that $r$ is generally quite a low integer,
and thus the overall computational complexity of both algorithms is equivalent.

 One main issue in the above formalism is the determination  of
 (the best estimation of) $r$ as this is indeed an unknown from a generic tensor data input. 
 This is the same as  determining the best number of principal components (PC) \cite{shlens2014tutorialPCA} for a given matrix.
 Different methods are proposed to determine the number of PC for the best reconstruction of the data. 
 Among these, we mention the scree plot \cite{cattell1966scree_testPCA} and the ratio methods \cite{abdi2010principalPCA}. The scree plot strategy displays the eigenvalues sorted in decreasing order and a threshold selection parameter for the vertical (eigenvalue) axis. 
 Then, it keeps the components with values above the threshold and removes the remaining. A good threshold is determined at the point where the eigenvalues drop significantly.\\
 \indent For a fixed slice $i$, the ratio method uses $\la_j^{(i)}/\sum_k \la_k^{(i)}$ where $\la_j^{(i)}$ is the eigenvalue related to the $j$-th PC. Here, one sorts the ratio  $(\la_j^{(i)}/\sum_k \la_k^{(i)})_j$ in decreasing order and selects the indices up to a significant drop. 
 For our problem, it turns out that
 both ideas are equivalent.

The following approach fixes the value of $r$.
 We have $m_1$ slices and the covariance matrix of each slice has $m_3$ eigenvectors. For the slice $i$, we denote by $n_i$ the number of eigenvectors selected by the scree plot method (equivalently by the ratio method)
 and  choose $r = \max_i n_i$. The implication of such a choice is as  follows: consider a $j$-th slice that has a number of selected vectors less than $r$,  i.e. $n_j < r$. 
Because the significant drop of the eigenvalues of the covariance matrix of the $j$-th slice happens at the position $n_j$,  the remaining $r - n_j>0$ eigenvectors are meaningless for the $j$-th slice. Thus, adding them will not affect the clustering process.  
 
\subsection{Cluster selection} For each mode, we have the similarity matrix $C'$ which records the similarity between all slices of this mode. To perform the clustering from these similarity matrices, we use  the spectral clustering (SC) \cite{BachLearningSpectralClustering} if the desired number of clusters in the three modes is given as an input $\left(k = (k_1, k_2, k_3)\right)$,  otherwise,  we use  the affinity propagation algorithm (AP) \cite{frey2007clustering}  that does not need such data.

\section{Algorithms}
\label{sec:algorithms}
In this section, we present the  two algorithms of MCAM for $3-$order tensors. As indicated in the previous section, these algorithms divide into two steps. The first step constructs the similarity matrix $C'$ using the expression (\ref{eq:Cg})
in one case, and  \eqref{eq:Cg2} in the other.  The second step uses the matrix $C'$ and applies a given clustering scheme (AP or SC) according to the user choice. 
The first option only needs the tensor data set as an input: we use the AP algorithm to detect the clusters.  
The algorithm \ref{algo:multiway_spectralI} 
\footnote[1]{The code is available at: https://github.com/ANDRIANTSIORY/MCAM} 
allows this option.
A second option requests the tensor data set and the desired number of clusters of each mode as input. In this case, we use the SC algorithm to find the different clusters.  The two algorithms easily adapt to such a situation. The output of the algorithm $(\cC_i)_{i\in[3]}$ represents respectively the clusters inside the three modes.

\begin{algorithm}[h]
		\caption{Multiway clustering via affinity matrix (MCAM-I)}
		\label{algo:multiway_spectralI}
\begin{algorithmic}[1]
         \Require $3$-order tensor $\cT\in\R^{m_1\times m_2\times m_3}$.
         \Ensure $(\mathcal{C}_1,\mathcal{C}_2,\mathcal{C}_3)$, the elements of the different clusters of the three modes. 
		\For{$j$ in $\{1,2,3\}$}
		    \For{ $i$ in $\{1,2,\cdots,m_j\}$}
				 \State Compute the spectral decomposition of $T_i^t T_i$.
			      \State Store the normalized eigenvectors $(\bx^{(i)}_1,\cdots, \bx^{(i)}_{k}, \cdots)$
				    \State Compute $n_i$ (scree-plot method).
			\EndFor
			\State $r \gets \max_in_i$
			\For{$k$ in $\{1,2,\cdots,r\}$}
			   \State {$ V_k \gets [\bx_k^{(1)} \bx_k^{(2)}\cdots \bx_k^{(m_j)}]$} 
			   \State {$C_{kk} \gets |V_k^t V_{k}|$}
			   \For{$k'$ in $\{1,2,\cdots,r\}\setminus\{k\}$}
			    \State {$C_{kk'} \gets |V_k^t V_{k'}|$ }
			   \EndFor
			\EndFor
			\State $C' \gets \frac{\la^2}{(\sum_{k=1}^r\la_{k})^2}\sum_{k=1}^{r}\sum_{k'=1}^{r} C_{kk'}$. 
			\State $\mathcal{C}_j\gets$ Affinity Propagation$(C')$.
		\EndFor
	\end{algorithmic}
\end{algorithm}

We present also a second version of the algorithm, that neglects cross similarities (algorithm \ref{algo:multiway_spectralII}). 
Therein, the lines 
8 to 11 replace the
lines 8 to 14 of the algorithm
\ref{algo:multiway_spectralI}.

\begin{algorithm}[h]
		\caption{Multiway clustering via affinity matrix (MCAM-II)}
		\label{algo:multiway_spectralII}
\begin{algorithmic}[1]
         \Require $3$-order tensor $\cT\in\R^{m_1\times m_2\times m_3}$.
         \Ensure $(\mathcal{C}_1,\mathcal{C}_2,\mathcal{C}_3)$, the elements of the different clusters of the three modes.
		\For{$j$ in $\{1,2,3\}$}
		    \For{ $i$ in $\{1,2,\cdots,m_j\}$}
				    \State Compute the spectral decomposition of $T_i^t T_i$.
				    \State Store the normalized eigenvectors
				    
				    $(\bx^{(i)}_1,\cdots, \bx^{(i)}_{k}, \cdots)$
				    \State Compute $n_i$ (scree-plot method).

			\EndFor
			\State $r \gets \max_in_i$
			\For{$k$ in $\{1,2,\cdots,r\}$}
			   \State $ V_k \gets [\bx_k^{(1)}\; \bx_k^{(2)}\cdots \bx_k^{(m_j)}]$ 
			   \State $C_{k} \gets |V_k^t V_{k}|$
			\EndFor
			\State $C' \gets \frac{\la^2}{\sum_{k'}\la_{k'}^2} \sum_{k=1}^{r} C_{kk}$. 
			\State $\mathcal{C}_j\gets$ Affinity Propagation$(C')$.
		\EndFor
	
	\end{algorithmic}
\end{algorithm}

\paragraph{Computational complexity} 
To  simplify  the evaluation, we request
  $m_i\in\Theta(n)$ for $ i= 1,2,3$. We fix $m_1=n$ and other dimensions  are  comparable  with $n$. The construction of the covariance matrix of each slice has a complexity $\cO(n^3)$ and the spectral decomposition has at most $\cO(n^3)$. Then, the computation of the covariance through all slices costs $\cO(n^4)$ complexity.  SC has a complexity  $\cO(n^3)$ and AP has a complexity at most  $\cO(n^4)$. So, the MCAM algorithm has a complexity class $\cO(n^4)$.

\newpage
\section{Experimentation}
\label{sec:experimentation}
In the following experiments, we apply the MCAM algorithms to synthetical datasets and one real dataset. For the synthetical datasets, the number of clusters in each mode  and the ground truth of elements are known. Hence, we evaluate the quality of the MCAM output by comparing it with the true cluster. We also compare the performance of our algorithms with three known clustering algorithms for multidimensional datasets: the  CP+k-means, the Tucker+k-means and the multiway clustering via tensor block  models (TBM). For the real data set, we evaluate the clustering  quality by computing the root means square error (RMSE) between the initial data and the estimated tensor generated from the clustering result.

\subsection{Synthetical datasets}
We compare the output of MCAM to the other algorithms with synthetic data. To do so,
we generate the  tensor datasets with the CP-decomposition.
We thus evaluate the clustering quality of the output of each algorithm by computing their clustering quality  criteria: the Adjusted Random Index (ARI) \cite{ARI} and the Normalized Mutual Information (NMI) \cite{NMI}. 
In all situations, we show that
MCAM performs with compelling results. 

We generate a $3$-order tensor 
dataset $\cT\in\R^{m_1\times m_2\times m_3}$ as defined in the equation \eqref{eq:CPdecomposition},
%
\begin{eqnarray}
    \cT &=& \sum_{j=1}^{d} \ga_j \bu_j\otimes \bv_j\otimes \bw_j + \cZ \\ \nonumber
    &=& [\Lambda; A, B, C] + \cZ
\end{eqnarray}
where $\Lambda = (\ga_1, \ga_2,\cdots,\ga_d)$, $A=[\bu_1 \cdots \bu_d]$, $B = [\bv_1\cdots\bv_d]$ and $C=[\bw_1\cdots\bw_d]$.\\

\begin{figure}[h!]
\centering
\includegraphics[width=11cm, height=5.5cm]{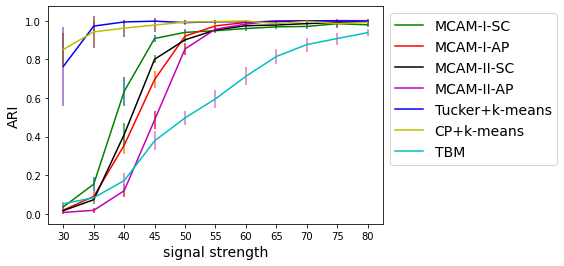}	 	
\caption{Comparison of the effectiveness between MCAM and the other multiway clustering  algorithms with a $n$ rank-one tensor dataset $(n=9)$.}	
\label{fig:tfs_data}
\end{figure}

The columns of the matrices $A$ are orthogonal, and so are those of $B$ and $C$ \cite{kolda2009tensor}. $\ga_j$ is a positive weight of the $j$-th rank-one tensor of the 
dataset, for all $j\in[d]$. $\cZ$ is a Gaussian noise tensor as defined in equation \eqref{eq:CPdecomposition}. 
We generate a tensor of size $m_1 = m_2 = m_3 = 100$ and each mode has the same number of clusters $c=9$. We fix the value of $(\ga_j)_{j\in[d]}$, by assuming that $\ga_1=\ga_2=\cdots=\ga_d = \gamma$. In each mode, we denote by $(J_i)_i$ the clusters. For each mode and for $j\in\{1,\cdots,d\}$, the column of A, B and C defined as $\bu_j(i)=1/\sqrt{|J_j|}$, $\bv_j(i)=1/\sqrt{|J_j|}$ and $\bw_j(i)=1/\sqrt{|J_j|}$ for $i\in J_j$ otherwise, $\bu_j(i)=\bv_j(i)=\bw_j(i)=0$. We vary the value of $\gamma$ from 30 to 80. For each value of $\gamma$, we repeat the experiment 10 times and present the mean and the standard deviation of the clustering quality index (ARI).  For each experiment, we regenerate the 
dataset and run the MCAM (I and II) with SP 
(coined in the figure MCAM-I/II-SP) or the MCAM (I and II) with AP 
(with index MCAM-I/II-AP) to the matrix $C'$. 
We also exhibit  the performance of  CP+k-means with rank decomposition $d$ equals the number of the clusters (i.e $d=c$), the performance of Tucker+k-means with rank decomposition equals to $(c,c,c)$ because we have $c$ clusters in each mode, as well as the performance of the TBM. For the convergence of the TBM, we use 50 iterations to update the membership matrices and the core tensor. 
Figure \ref{fig:tfs_data} 
illustrates the performances of the different algorithms.

Figure \ref{fig:tfs_data} shows that the MCAM-AP  (I and II) performs well when the value of $\gamma$ increases. 
We have a signal tensor with 9 clusters and they cover more than 95\% of the entries of the tensor dataset. 
Naturally, 
this improves the 
clustering results of the MCAM-AP method.
 Furthermore, for the present data, we see that MCAM-I performs better than MCAM-II either with SC or with AP tools.

The graphics also reveal  that for a small value of $\gamma$  CP+k-means and Tucker+k-means have the best performance. Remember that the rank decomposition in these methods is  equal to the number of clusters in each mode, and this favors them  
versus the other methods. 
(The following paragraph discusses this fact). From $\gamma=55$, the MCAM has the quality of the CP+k-means and  Tucker+k-means methods. However, good results from the TBM require a higher value of $\gamma$ (more than 80) to detect all the expected clusters in this dataset model.

For $\gamma\ge 55$,  MCAM-AP (I and II) recovers the expected number of clusters inside the data, i.e. $c = 9$ with their respective elements, without taking the number of clusters as an input, as opposed to the other algorithms. By performing 10 times the experiments, for both MCAM-I and -II methods, the most frequent value of $r$ 
is determined at $2$. This  means that the MCAM algorithm only needs the two largest eigenvectors of each slice to build the similarity matrix  (equation \eqref{eq:Cg}) and cluster the data. 

\begin{figure}[h!]
\centering	 	
\includegraphics[width=10cm, height=6cm]{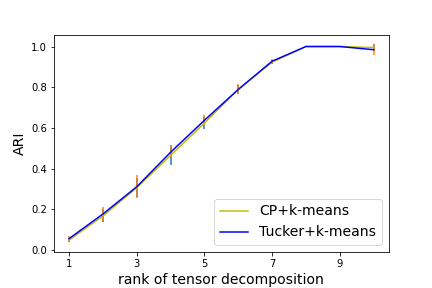}	 
\caption{The two figures represent the quality of the clustering output according to the rank of the tensor decomposition, CP decomposition and Tucker decomposition, with $\gamma = 55$.} 	 	
\label{fig:rankdec}
\end{figure}

Another aspect needs emphasis: 
the CP+k-means or Tucker+k-means perform a good clustering only with a high enough $d$ of rank-one decomposition (see figure \ref{fig:rankdec}). Thus, the higher $d$ is, the better
CP+k-means or Tucker+k-means
will behave. On the other hand, the MCAM does not require this rank decomposition $d$ but generates the number $r$, which we will call the "effective clustering dimension" (ECD). It is an interesting question to ask whether the access of $d$ rank-one decomposition has a greater cost than that of the ECD $r$ (just linear in the dimension of slices), in which 
case, MCAM would be even more efficient than the other algorithms.
We observe that even at $r \simeq 2$ (for $\gamma\ge55$), MCAM delivers qualitatively excellent results.

\begin{figure}[h]
\centering
\includegraphics[width=10cm,
height=6cm]{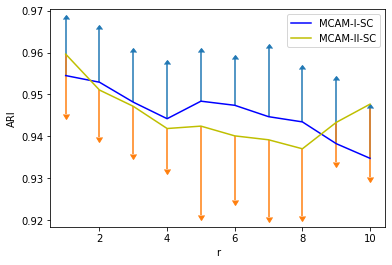}	 
\caption{The clustering quality (ARI) for $r$ varying from 1 to 10 and $\gamma = 55$.}	 
\label{fig:boxplot60}
\end{figure}

To study the dependence of the MCAM algorithm on $r$, we run the algorithms \ref{algo:multiway_spectralI} and  \ref{algo:multiway_spectralII} by varying the value of $r$ from 1 to 10 (see equations \eqref{eq:Cg} and \eqref{eq:Cg2}) by constructing the similarity matrix associated with the dataset with signal strength $\gamma=55$.
For each $r$, we run the algorithm MCAM-SC $10$ times. Therefore, we have 10 quality indices of the output (ARI) and we compute their standard deviation and their mean.  The figure \ref{fig:boxplot60} 
show  the 10 means of the quality indices
of the output 
from the two different algorithms of MCAM. 
We observe that the values are very close to 1. Hence, 
even at $r=1$, the MCAM can provide a good output
(ARI $\ge 0.95$ for MCAM-I and MCAM-II).
This preliminary analysis suggests that MCAM can be efficient up to a fixed and small integer $r$.

\subsection{Real Data}

\begin{figure}[h!]

\centering	 	
\includegraphics[width=10cm, height=6cm]{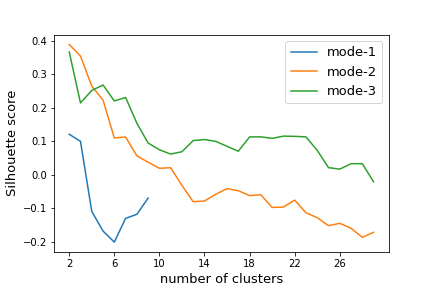}	  	
\caption{The silhouette score of the mode-1,  of the mode-2, and of the mode-3.} 	 	
\label{fig:silhouette}
\end{figure}

Let us address the case of the  real dataset. 
 We run the MCAM-I and -II on the flow injection analysis (FIA) dataset \cite{fiadata}. This dataset has a size of $12$ (samples) $\times 100$ (wavelengths) $\times 89$ (times). For each mode of this tensor data, we do not have any information about the exact number of clusters. 
 For MCAM-SC, CP+k-means, 
 Tucker+k-means and TBM,
 we apply the silhouette score \cite{SilhouettesRousseeuw1987} 
 to the similarity matrix $C'$ to detect the number of clusters in each mode.  We evaluate a range of the possible number of clusters $c$ in each mode, (the result is plotted in figure \ref{fig:silhouette}.
 For the CP+k-means algorithm, the number of rank-one  decompositions of the tensor is equal to the maximum among the number of the clusters in the three modes ($d= \max(c_1,c_2,c_3)$). For the Tucker+k-means algorithm, the rank decompositions of the tensor is equal to the number of the clusters in each mode (${\rm rank} = (c_1, c_2, c_3)$).
 The RMSE \cite{ChaiRMSEMAE2014} 
 will serve as an index of comparison of performance
 and quality between the different algorithms.

For mode-$1$, we vary the number of clusters $c$ from $2$ to $6$. For  mode-$2$ and mode-$3$, we variate the number of clusters from $2$ to $30$. Using the silhouette score, we infer that there are $2$ clusters in each mode and therefore we have $2\times 2\times 2 = 8$ sub-tensors in the dataset (see figure \ref{fig:silhouette}). The quality of the output of the algorithms is shown in table \ref{tab:rmse_silhouette}.
Once again, the MCAM outperforms its two rival methods CP+k-means and Tucker+k-means by having a smaller RMSE. We assume to reach the convergence of TBM after 50 iterations. The RMSE tells us that TBM has the best clustering result.

\begin{table}[h!]
\caption{The mean of RMSE of all sub-tensors for the number of clusters selected by the Silhouette method.}
\label{tab:rmse_silhouette}
\vskip 0.15in
\begin{center}
\begin{small}
\begin{sc}
\begin{tabular}{|c|c|}
\hline
\textbf{Method} & \textbf{RMSE} \\
\hline
\textbf{MCAM-I-AP} & $0.0685$ \\
\hline
\textbf{MCAM-I-SC} & $0.0831$ \\
\hline
\textbf{MCAM-II-AP} & $0.0685$ \\
\hline
\textbf{MCAM-II-SC} & $0.0831$ \\
\hline
\textbf{CP+k-means} & $0.0843$ \\
\hline
\textbf{Tucker+k-means} & $0.0842 $ \\
\hline
\textbf{TBM} & $0.0661$   \\
\hline
\end{tabular}
\end{sc}
\end{small}
\end{center}
\vskip -0.1in
\end{table}

It is noteworthy that
the MCAM-AP (I and II) delivers directly a
different number of clusters for each mode as $(4,6,9)$. Hence, it is expected that it has a low RMSE mean compared to the other methods.
At this moment, We see that the TBM has the best clustering result among the four algorithms which take the number of clusters as an input.  We also realize that  MCAM provides a better clustering than produced by the CP+k-means and Tucker+k-means.

\section{Conclusion}
\label{sec:conclusion}
Multiway clustering aims at partitioning all entries of a tensor into  pairwise disjoint  sub-tensors that define the clusters.
The MCAM introduced in this paper is a new multiway clustering algorithm 
 based on affinity matrices that record the data similarity between the tensor slices. 
In our procedure, the cluster selection in each mode proceeds independently from the other modes. For each mode, the determination of the cluster is divided into two parts: the first part is the construction of the similarity matrix $C'$ and the second part is the selection of the elements of each cluster.
 We have proposed two algorithm versions (MCAM-I and MCAM-II), both using two matrix clustering  subroutines (SC and AP).
Based on the affinity propagation method, one subroutine (AP) makes the method more generic than several clustering algorithms (MCAM-AP), 
as it does not necessarily take  the number of clusters as an input. 
The second method 
MCAM-SC (for versions -I and -II) uses  spectral clustering but requires the number of clusters
as an input. 
The MCAM procedure is evaluated on a standard synthetic  and one real dataset providing  excellent results. 
The clustering obtained from the MCAM is compared to the upshots of three other algorithms (CP+k-means, Tucker+k-means and TBM). In all experiments with a  synthetic dataset, the MCAM shows a high performance  compared to the other methods. 
Finally, the strong point of MCAM  is that the clustering selection proceeds via an affinity matrix to which other methods might apply. Translating the tensor data into matrix data could have a great advantage on the scalability of the algorithm. This deserves further investigation.  Another extension of this work should deepen the understanding of the similarity matrix of the MCAM method.

%
%
%
\bibliographystyle{splncs04}
\bibliography{myfile}
\end{document}